\pdfoutput=1

\documentclass[11pt]{article}

\usepackage[]{naacl2021}

\usepackage{times}
\usepackage{latexsym}

\usepackage[T1]{fontenc}

\usepackage[utf8]{inputenc}

\usepackage{microtype}

\title{Self-supervised Text-to-SQL Learning \\With Header Alignment Training}

\author{Donggyu Kim$^1$ \quad Seanie Lee$^{2}$ \\  
	Kookmin Bank$^1$,  KAIST$^2$, South Korea\\
	\texttt{donggyukimc@gmail.com, lsnfamily02@kaist.ac.kr}}
	
\renewcommand\footnotemark{}
\date{}

\usepackage{tabto}
\usepackage{pgfplots}
\usepgfplotslibrary{groupplots}
\usepackage{times}
\usepackage{latexsym}
\usepackage{mathtools}
\usepackage{colonequals}
\usepackage{comment}
\usepackage{multirow}
\usepackage{anyfontsize}
\usepackage{tabu}
\usepackage{booktabs}
\usepackage{amsmath,amssymb}
\usepackage{verbatim}
\usepackage{bm}
\usepackage{pgfplots}
\usepackage{enumitem}

\usepackage[skip=2pt]{caption} 
\usepackage{float}
\usepackage{tikz}
\usepackage{mathtools}
\usepackage{xcolor,soul}
\usepackage{scrextend}
\usepackage{float}
\usepackage{xcolor}

\usepackage{listings}

\DeclareFixedFont{\ttb}{T1}{txtt}{bx}{n}{12} 
\DeclareFixedFont{\ttm}{T1}{txtt}{m}{n}{12}  

\usepackage{color}
\definecolor{deepblue}{rgb}{0,0,0.5}
\definecolor{deepred}{rgb}{0.6,0,0}
\definecolor{deepgreen}{rgb}{0,0.5,0}

\usepackage{listings}

\newcommand\pythonstyle{\lstset{
language=Python,
basicstyle=\ttm,
otherkeywords={self},             
keywordstyle=\ttb\color{deepblue},
emph={MyClass,__init__},          
emphstyle=\ttb\color{deepred},    
stringstyle=\color{deepgreen},
frame=tb,                         
showstringspaces=false,           %
caption={Python code for TaBERT},captionpos=b,label=code
}}

\lstnewenvironment{python}[1][]
{
\pythonstyle
\lstset{#1}
}
{}

\begin{document}
\maketitle

\begin{abstract}
Since we can leverage a large amount of unlabeled data without any human supervision to train a model and transfer the knowledge to target tasks, self-supervised learning is a de-facto component for the recent success of deep learning in various fields. However, in many cases, there is a discrepancy between a self-supervised learning objective and a task-specific objective. In order to tackle such discrepancy in Text-to-SQL task, we propose a novel self-supervised learning framework. We utilize the task-specific properties of Text-to-SQL task and the underlying structures of table contents to train the models to learn useful knowledge of the \textit{header-column} alignment task from unlabeled table data. We are able to transfer the knowledge to the supervised Text-to-SQL training with annotated samples, so that the model can leverage the knowledge to better perform the \textit{header-span} alignment task to predict SQL statements. Experimental results show that our self-supervised learning framework significantly improves the performance of the existing strong BERT based models without using large external corpora. In particular, our method is effective for training the model with scarce labeled data. The source code of this work is available in GitHub.~\footnote{We will release the codes, pretrained models upon acceptance of our paper.} 
\end{abstract}

\section{Introduction}

\begin{figure*}[t]
    \begin{center}
    \includegraphics[scale=0.48]{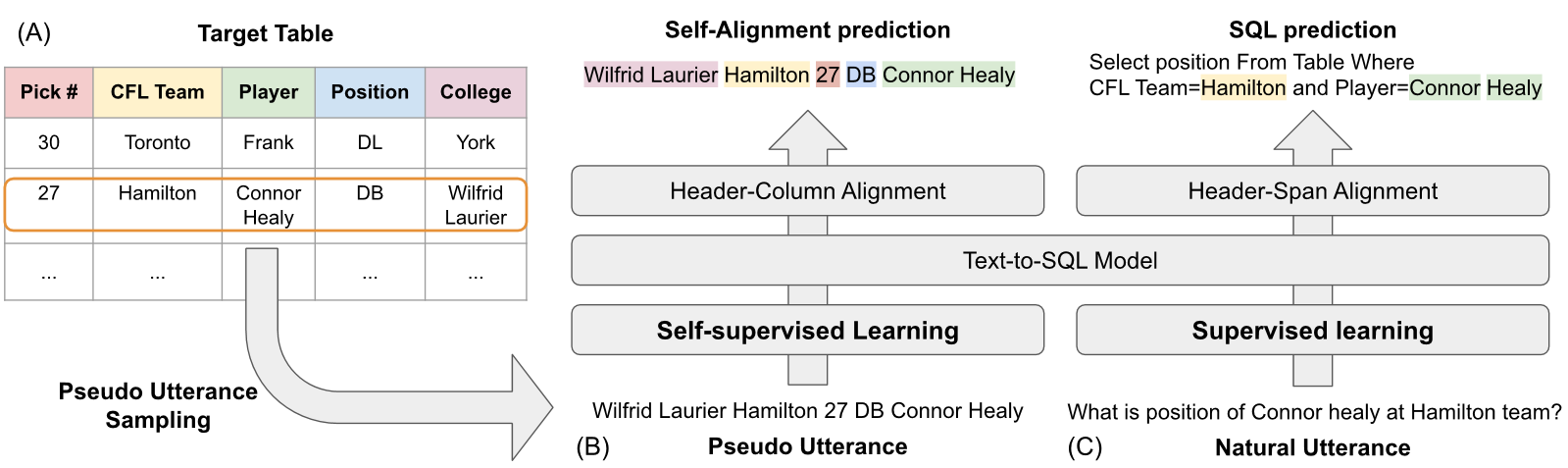}
	\vspace{0.05in}
	\caption{(A) Given a table, (B) we train a model to align table headers with \textit{pseudo utterance} in self-supervised learning, (C) and further train the model with supervised learning for Text-to-SQL. Both training procedures share same Text-to-SQL model.}
	\vspace{-0.2in}
    \label{concept-fig}
    \end{center}
\end{figure*}

In recent works on Natural Language Understanding (NLU), it is common practice to utilize self-supervised learning approaches~\cite{peters2018deep, devlin2019bert, Lewis2019BARTDS, raffel2019exploring} for training a model on a large amount of unlabeled corpora and fintune it with a task-specific objective. The self-supervised learning becomes more crucial in the setting where there are only few human annotated training data. Since the knowledge that the model learns from self-supervised training can be transferred to target tasks such as GLUE \citep{Wang2018GLUEAM} with small amount of labeled data, the model pretrained with self-supervised learning is more data-efficient and robust to unseen domain \citep{Hendrycks2020PretrainedTI}.

The key point to build good self-supervised learning scheme is to design a learning objective satisfying following conditions. 1) We can sample a pair of input and its  corresponding label without human annotation. 2) models trained with such unlabeled data should understand the underlying structure of target data (e.g. pixel relations in images and word relations in corpus) so that the knowledge is helpful for finetuning the model with the task-specific objectives. The most well-known example is Masked Language Modeling (MLM) \citep{devlin2019bert, liu2019roberta, clark2020electra}, which trains a model to predict randomly masked words. With such self-supervision, models  can learn the contextual knowledge from corpora, such as phrases or relational knowledge \cite{lm-as-kb}. Those knowledge can be leveraged to improve the performances of target NLU tasks. 

In Text-to-SQL task \cite{Shi2018IncSQLTI, seq2sql}, we are able to retrieve an answer for a given utterance only after executing corresponding formal SQL statements. Therefore, the supervised learning for Text-to-SQL typically requires annotated training samples --- a set of triplets consisting of a natural utterance, a table, and the corresponding SQL statement as shown in \ref{concept-fig}-(C), which have to be annotated by those with expert knowledge of SQL. Due to such constraints, it is difficult to deploy Text-to-SQL models in new unseen domain. Even though there are data-efficient approaches like weakly-supervised learning \citep{liang2018memory, min2019discrete}, they largely underperfom models trained with annotated data. Recently, some works \citep{hwang2019comprehensive, rat-sql, choi2020ryansql} leverage the pretrained models for Text-to-SQL task and  significantly improve the performance. However, there is still a discrepancy between MLM objective and Text-to-SQL objective which requires a reasoning over table data for a given utterance.  

To overcome the shortcoming of data-inefficiency in supervised Text-to-SQL training and discrepancy between self-supervised learning and finetuning, we propose a novel self-supervised learning framework for Text-to-SQL. We leverage the property of Text-to-SQL task where a model predicts SQL statements by aligning the given table headers with text spans in the utterances. Specifically, we train a model to extract the start and end positions of the text span, which is an entity corresponding to a certain table header, from the \textit{pseudo utterance}. We define the pseudo utterance as a concatenation of entities from columns sampled from unlabeled table contents. For example, as shown in Figure \ref{concept-fig}-(B), given the table header and pseudo utterance --- concatenation of the entities from the second rows, the model is trained to predict start and end position of ``Hamilton" for the header ``CFL Team". With the proposed self-supervised learning, we can enforce the model to reason over tables with the pseudo-utterance, which is an essential component for Text-to-SQL.  As a result, the model can learn knowledge on \textit{header-column} alignment task and directly leverage the knowledge for the target \textit{header-span} alignment task of supervised Text-to-SQL learning. 

Our proposed method is model-agnostic and generally applicable to any Text-to-SQL models by training the model with our self-supervised method and further fintuning it with all Text-to-SQL objectives. 
We show that our method significantly improves the performances of existing baselines on the two well-known Text-to-SQL benchmark datasets --- WikiSQL \citep{seq2sql} and Spider \citep{yu2018spider} without any additional parameters or extra training data other than the given dataset, contrary to the previous works \citep{yin20acl, tapas} which utilize enormous amount of external corpora for self-supervised learning.
Moreover, we empirically validate that our proposed method achieve much larger gains in low-resources settings where there are few annotated training samples.

To sum up, our contribution is threefold:
\begin{itemize}[itemsep=0.8mm, parsep=0pt, leftmargin=*]
    \item We propose a novel method of self-supervised learning for any Text-to-SQL models, where the model is trained to learn the alignment between \textit{pseudo utterances} and table headers.
    
    \item Without any extra large amount of table data, our proposed method using only the  tables from the given dataset improves the performance of strong baselines (BERT-base/large) on two Text-to-SQL benchmark datasets ---   WikiSQL and Spider.
    
    \item We empirically validate that our framework is effective for low-resource environment where there are few labeled data. 
\end{itemize}

\section{Related Work}
\noindent
\paragraph{Fully-Supervised Text-to-SQL}  
A model is trained to predict SQL statements from given utterances, and tables with labeled data. The self-supervised method usually adopts either slot-filling \citep{hwang2019comprehensive, yu2018syntaxsqlnet} or sequence-to-sequence learning with encoder-decoder architecture \cite{zhang2019editing, sun-etal-2018-semantic, seq2sql, Lin2019GrammarbasedNT}. It requires a set of triplets consisting of an utterance, a table, and an annotated SQL statement as labeled data to train the model with supervised learning.

\noindent
\paragraph{Weakly-supervised Text-to-SQL}  This approach has been recently spotlighted since it alleviates the burden of annotating the SQL statements. \citet{Agarwal2019LearningTG} utilize a set of triplets consisting of utterances, tables, and the answers of utterance (i.e., the execution results of SQL statements) rather than annotated SQL statement.  They train a model with reinforcement learning in order to obtain a training signal from the execution of SQL. 
\citet{tapas} formulates Text-to-SQL as a question answering over tables without any generation of logical forms by extending BERT architecture to encode tables as inputs. In spite of such advantage of weak-supervision, the performance of these models are largely behind those with fully-supervised learning.

\noindent
\paragraph{Self-supervised Learning} Regardless of domains and tasks, self-supervised learning is a popular approach in recent deep learning works because of its data-efficiency and effectiveness. Typically, self-supervised learning uses user-defined training objectives in order to learn the knowledge on underlying structures of target data, which will be helpful for specific tasks with the target data. There are several works \cite{Chen2020ASF, fixmatch} for computer vision tasks.
Masked language modeling (MLM) proposed by \citet{devlin2019bert} is the most representative for self-supervised learning in natural language processing. It utilizes two self-supervised objectives, Masked Token Prediction (MTP) and Next Sentence Prediction (NSP). MTP  randomly masks words from contexts and predicts the masked tokens based on the surrounding contexts. On the other hand, NSP predicts whether two contiguous text segments are coherent or not. The model can learn the fruitful knowledge on natural language with such self-supervision. There have been further progress \citep{clark2020electra, liu2019roberta} and variations \citep{Lewis2019BARTDS} on MLM.

Recent works \citep{tapas, yin20acl} propose self-supervised learning method with large scale semi-structured table corpora. In contrast, we propose an alternative self-supervised learning framework where a model benefits from a much smaller amount of table data. In our method, the model is trained to learn alignment between table headers and columns, which we call header-column alignment. Since the header-column alignment task is closely related to the header-span alignment task, which is essential ability for the Text-to-SQL task, the model is able to directly learn the knowledge of Text-to-SQL with our proposed method.

\section{Method}
\subsection{Background: Text-to-SQL}
The goal of Text-to-SQL is to convert a given natural language utterance $\mathbf{u} = (u_1, \ldots, u_L)$, consisting of $L$ tokens, to SQL statement with formal structures, which are typically composed of \textit{select-clause} and \textit{where-clause} parts. To fill in the select-clause, the model predicts a header from a given set of table headers (i.e. schema) $\mathbf{h} = (h_1, \ldots, h_M)$ and a proper aggregation (e.g. sum, count, avg). In addition, for where-clause, the model predicts corresponding headers for conditions, operations (e.g. $>, <,$ and $=$), and text spans (i.e., sub-sequences of given utterance) for condition values.

In order to construct those sub-components of SQL statements, one essential ability of the model is to learn an alignment between the given table headers with the text spans of values in the utterance. As the example in Figure \ref{concept-fig}-(C), the model should  align the spans in the utterance "Connor Healy" and "Hamilton team" with its corresponding table headers "Player" and "CFL Team", respectively.  Therefore, we train the neural network model parameterized with $\phi$ to maximize the following conditional log likelihood of the correct
\textit{start} and \textit{end} position of spans in the given utterance $\mathbf{u}.$
\begin{align}
\begin{split}
    \max_\phi \sum\limits_{i=1}^M \big\{&\log p_\phi(u_{{start}_i}|h_i, \mathbf{u})\\
    +&\log p_\phi(u_{{end}_i}|h_i, \mathbf{u}) \big\}
    \label{eq:1}
\end{split}
\end{align}

\subsection{Self-Supervised Learning}
\label{2_method}
Since tables are structured data consisting of the columns and their headers, the contents stored in columns can be aligned with their corresponding headers without any labels for SQL statements. For example, as shown in Figure \ref{concept-fig}-(A), we sample the second row of the table and concatenate its entities with a random order as  ``Wilfrid Laurier Hamilton 27 DB Connor Healy" ---  we call it as pseudo utterance. Given a pair of  headers ``Pick \#", ``CFL Team", ``Player", ``Position",``College" and its corresponding spans in the pseudo utterance, we can train the model to predict start and end position spans of each entity (e.g., ``Connor Healy") from the pseudo utterance for its corresponding header (e.g., ``Player") without any human annotation, as shown in Figure \ref{concept-fig}-(B).

If the given utterance is answerable from the certain table, the pseudo utterance sampled from the same table include the target text spans of the table which are to be aligned with the table header to construct a proper SQL statement for the given utterance. For instance, as shown in Figure \ref{concept-fig}-C), the valid target text spans of utterance, ``Hamilton" and ``Connor Healy" are included in the pseudo utterance, ``Wilfrid Laurier Hamilton 27 DB Connor Healy" from Figure \ref{concept-fig}-(B). Therefore, training the model to align the table headers with text span from pseudo utterance is a valid training signal to obtain the knowledge about the underlying structures of tables. Furthermore, there is no significant discrepancy between training the model to align the header with the utterance, which we call \textit{header-span alignment}, and align the header with the pseudo utterance, named as \textit{header-column alignment} because two training procedures can share same model parameters. As a result, the knowledge that the model learns from the header-column alignment task will be helpful for further supervised training for the header-span alignment task.

Based on the intuition, we propose a novel self-supervised learning framework for Text-to-SQL, which is to align headers with its corresponding entities from columns of the given table, as follows. We construct a pseudo utterance by sampling a row of the given table and concatenate the entities from each column of the sampled row. Then we maximize the log likelihood of entity spans for corresponding headers of the pseudo-utterance. 
\begin{table*}
	\centering
	\begin{tabular}{lcccccc}
	    \midrule[1.0pt]
	    {\textbf{Dataset}} & \multicolumn{3}{c}{\bf WikiSQL} & {} & \multicolumn{2}{c}{\bf Spider}\\
	    \midrule[0.4pt]
		{\textbf{Split}} &{\textbf{Train}} &{\textbf{Dev}} &{\textbf{Test}} & {} &{\textbf{Train}} &{\textbf{Dev}} \\
		\midrule[0.4pt]
		{\# of Samples} & {56,355} & {8,421} & {15,878} & {} & {8,495} & {1,034} \\
		{\# of Schema}  & {-}  & {-}  & {-} & {}  & {146} & {20} \\
		{\# of Tables} & {18,585} & {2,716} & {5,230} & {} & {795} & {81} \\
		{\# of Rows} & {317,769} & {44,761} & {96,290} & {}  & {59,208} & {5,481} \\
		\midrule[1.0pt]
	\end{tabular}
	\small\caption{Statistics of data splits in Text-to-SQL benchmark datasets, WikiSQL and Spider. It shows the number of annotated samples with utterance-SQL statements, the number of Tables, and the number of rows in all tables. 
	}
	\vspace{-0.2in}
	\label{table_dataset}
\end{table*}

\noindent
\paragraph{Pseudo Utterance Sampling}  To sample pseudo utterances from table contents, we propose three variants of sampling methods as follows:
\vspace{-0.1in}
\begin{enumerate}[itemsep=0.8mm, parsep=0pt, leftmargin=*]
    \item concatenate entities from the same row of table with fixed order
    \item concatenate entities from the same row of table with random order
    \item sample entities from columns of different tables and concatenate them with random order
\end{enumerate}
For the first method, the model would easily predict the alignments based on the sequential order since entities from each column is sampled and concatenated without any manipulations which include shuffling or injecting a noise. Therefore, it is hard to obtain meaningful learning signal from the \textit{pseudo utterance} sampled in this manner. For the second method, however, the order of the entities from each column is shuffled so that the model should align the entities with the header based on their contextual reasoning over the underlying structure of the table. Meanwhile, the third sampling method aims to sample more challenging \textit{pseudo utterance}  because we sample some entities from the target table and the others from another ones. The model should consider negative cases where some of headers do not have their corresponding entities in the sampled pseudo utterance. However, we empirically find out that there is no improvement from the third method compared to the others. Thus, we choose the second sampling method for our self-supervised training.

\noindent
\paragraph{Objective Function}  
Given a set of $N$ unlabeled pairs of the header and pseudo utterance $\{\mathbf{c}^{(i)}, \mathbf{h}^{(i)} \}_{i=1}^N$, where $\mathbf{c}^{(i)} = (c^{(i)}_1, \ldots, c^{(i)}_K)$ is a pseudo utterance consisting of $K$ tokens, we maximize  the following self-supervised objective function with respect to the parameter $\phi$ concerning header-column alignment task.
\begin{align}
    \begin{split}
    \mathcal{L}_{self} = \frac{1}{N}\sum\limits_{i=1}^{N}\sum\limits_{j=1}^{M} \big[\log p_{\phi}(\textit{c}_{start_{j}}^{(i)} | \mathbf{\textit{h}}^{(i)}_j, \mathbf{c}^{(i)})  \\
    + \log p_{\phi}(\textit{c}_{end_{j}}^{(i)} | \mathbf{\textit{h}}^{(i)}_j, \mathbf{c}^{(i)}) \big]
    \end{split}
\vspace{-0.11in}
\end{align}

\subsection{Supervised Learning}
\label{3_method}
After the pretraining with the self-supervised objective function, we finetune the model to maximize the supervised Text-to-SQL objective function with a set of $N'$ triplets, $\{\mathbf{s}^{(i)}, \mathbf{u}^{(i)}, \mathbf{h}^{(i)} \}_{i=1}^{N'}$ where $\mathbf{s}^{(i)}$ denotes an annotated SQL statements and each $\mathbf{s}^{(i)}$ consists of $T_i$ tokens. Note that the parameters $\theta$ is a set of all the trainable parameters of the model including $\phi$.
\begin{align}
    \begin{split}
    \mathcal{L}_{SQL} = \frac{1}{N'}\sum\limits_{i=1}^{N'}& \log p_{\theta}(\mathbf{s}^{(i)} | \mathbf{u}^{(i)}, \mathbf{h}^{(i)})
    \end{split}
\end{align}
Please note that our framework is generally applicable to existing models for Text-to-SQL by simply performing self-supervised training before supervised training. Another advantage of our framework is that the model learns knowledge from the table contents in training time and utilize the knowledge without accessing the contents in the test time. It is much more scalable to employ the contents of table, compared to the aforementioned existing methods using the table contents in  the test time, which is computationally prohibitive and slows down the prediction.

\section{Experiments}
\label{experiments}

\begin{table*}
	\centering
	\begin{tabular}{lccc}
		\toprule[1pt]
		
		{} &\multicolumn{2}{c}{\textbf{WikiSQL}} &{\textbf{Spider}} \\
		\midrule[0.4pt]
		\textbf{Method} &{\textbf{dev (L/E)}} &{\textbf{test (L/E)}} &{\textbf{EM}}\\

		\midrule[0.4pt]
		{BERT-Base} & {81.65 / 89.45} & {81.35 / 89.39} & {43.90}\\
		{Base-MLM} & {71.93 / 84.15} & {71.73 / 84.51} & {43.60}\\
		{Base-TaBERT} & {82.91 / 90.33} & {82.66 / 90.21} & {44.96}\\
		\textbf{Base-Self (Ours)} & {\textbf{83.66} / \textbf{90.68}} & {\textbf{83.61} / \textbf{90.84}} & {\textbf{45.25}}\\

		\midrule[0.4pt]
		{BERT-Large} & {82.42 / 90.14} & {82.34 / 90.21} & {56.10}\\
		{Large-MLM} & {83.67 / 90.61} & {83.38 / 90.55} & {56.69}\\
		{Large-TaBERT} & {84.11 / 91.05} & {84.20 / 91.10} & {52.42}\\
		\textbf{Large-Self (Ours) } & {\textbf{84.87} / \textbf{91.34}} & {\textbf{84.57} / \textbf{91.21}} & {\textbf{58.43}}\\
		
		\midrule[1pt]
	\end{tabular}
	\caption{Experimental results on \textbf{WikiSQL} and \textbf{Spider}. For WikiSQL, we report logical-form and execution accuracy, denoted as \textbf{L} and \textbf{E}, respectively. For Spider, we report Exact-Set Match with value, denoted as \textbf{EM}.}
	\vspace{-0.1in}
	\label{main-table}
\end{table*}
\subsection{Dataset}
We use two well-known Text-to-SQL benchmark datasets, WikiSQL\footnote{https://github.com/salesforce/WikiSQL} \citep{seq2sql} and Spider\footnote{https://github.com/taoyds/spider} \citep{yu2018spider}, to evaluate the effectiveness of our self-supervised Text-to-SQL learning approach. Table \ref{table_dataset} shows the overall statistics of datasets, the number of samples, tables, and rows. The number of tables from WikiSQL is much larger than the other Text-to-SQL datasets \citep{atis, zettlemoyer2005learning, iyer-etal-2017-learning} including Spider. On the other hand, Spider has the smaller number of tables, but it contains diverse and cross domain tables. Moreover, SQL statements of Spider consist of much complex components such as Group-by and Having clauses. 

\subsection{Implementation detail}
\paragraph{Baseline}
We compare our method against relevant baselines.
\begin{enumerate}[itemsep=0.8mm, parsep=0pt, leftmargin=*]

    \item \textbf{BERT-Base/Large:} For WikiSQL dataset, this is the  model of which backbone network is pretrained BERT base or large \cite{devlin2019bert}, proposed by \citet{hwang2019comprehensive}, and enhanced with Execution-Guided (EG) decoding method \citep{Wang2018RobustTG}. For Spider dataset, it is the variant of RyanSQL model \cite{choi2020ryansql} modified to generate complete SQL statement with value.\footnote{We choose the model because it is the end-to-end model and  generates SQL statements without accessing the contents of tables,  unlike other works \citep{Guo2019TowardsCT, rat-sql} that use hand-crafted features or require to access the tables during the test time.} 
    
    \item \textbf{Base/Large-MLM:} This is \textbf{BERT-Base/Large} model further pretrained  with Masked language modeling (MLM) objective on unlabeled table data, which is proposed by \citet{Gururangan2020DontSP}. The model is trained to predict the masked tokens from the concatenated sequences of headers or entities from the headers.
    
    \item \textbf{Base/Large-TaBERT:} It is the variant of BERT model which is pretrained on a large corpus of 26 million tables and their English contexts, proposed by \citet{yin20acl}. We replace the backbone network of \textbf{BERT-Base/Large} with the pretrained TaBERT model provided from Github page.  \footnote{https://github.com/facebookresearch/TaBERT} Please see \textbf{Supplementary File} for more implementation detail. 

    \item \textbf{Base/Large-Self:}  This is the same model as \textbf{BERT-Base/Large} but trained with our self-supervised learning method and finetuned with the supervised Text-to-SQL objective.
\end{enumerate}

\paragraph{Prediction with Value} Most of existing works on Spider \citep{Guo2019TowardsCT, rat-sql, choi2020ryansql, yu2018spider, yin20acl} report performances on Exact-Set Match without value. It indicates the models do not generate complete SQL statements with actual values for SQL conditions (e.g. "James" in SELECT age FROM table WHERE name="James"). Since such outputs are meaningless for real world applications, we report Exact-Set Match with value for our experiments.

\paragraph{Pseudo utterance sampling} In the case of WikiSQL, each training instance is associated with only a single table. For Spider, however, there are multiple tables for a single schema. To incorporate the knowledge about dependencies among the multiple tables, we concatenate all the tables for each corresponding schema as a single table and exclude redundant or duplicate columns for foreigner keys such as ID and Code columns during pseudo utterance sampling procedure described in Section \ref{2_method}. We also truncate the columns of number types, which are not useful to train the semantic relationship between columns and values. Note that, during the self-supervised training of the models for WikiSQL and Spider, we only use unlabeled table data from their own datasets in order to show the  effectiveness of our method in low resource environments where there are only few tables.

\paragraph{Experimental Setup}  We train and evaluate the models with the same hyper-parameters as described in the original works \citep{hwang2019comprehensive, choi2020ryansql}. For WikiSQL model, we set the batch size as 32 and beam size as 8, and use Adam \citep{kingma2014adam} optimizer with learning rate $10^{-5}$. We train the model for 30 epochs and report the performances of the model which shows the best overall performances on both development (dev) set and test set. For the model trained on Spider dataset, we measure the best performances of models on dev set. Following  \citet{choi2020ryansql}, we train the models until they do not show further improvements over 20 training epochs. We also follow  the same hyper-parameter setting where batch size is 8 and learning rate is $10^{-5}$.

As described in \ref{3_method}, before fine-tuning the models on labeled samples, we perform our self-supervised learning up to 3 epochs. With a single V-100 GPU machine, it takes 3 hours to train Base models and 6 hours to train Large models for our proposed self-supervised learning with the table data in WikiSQL. For Spider dataset, it takes less than a hour to train both Base and Large models with our proposed self-supervised objective function because the dataset contains much smaller amounts of table data.

\subsection{Experimental Results}
We report the experimental results on the baseline models and the models trained with our proposed self-supervised learning objective in Table \ref{main-table}. 
We report Logical-form (\textbf{L}) and Execution (\textbf{E}) accuracy for WikiSQL and Exact-Set Match (\textbf{EM}) with values for Spider.

For WikiSQL, regardless of model capacity, our method consistently outperform the baseline models, except TaBERT, with large margins. To be specific, it significantly improves BERT-Large, which is  +2.45\% and +2.23\% logical-form accuracy improvements for dev and test set, respectively.  For execution accuracy, our method improves it by +1.2\% for dev set and +1.0\% for test set. In contrary to ours, MLM marginally improves or even degrades the performance. Moreover, please note that our models without any additional tables show slightly better performance than TaBERT which is pretrained on 26 millions of tables and their English context, which is about \textbf{11,837} times larger than the number of tables from WikiSQL. 

For Spider dataset, with using the much smaller number of tables for self-supervised learning, our proposed self-supervised learning framework consistently improves the performance of all the baseline models. We speculate the reason why TaBERT-Large model significantly underperforms the other baselines is disability of the model from the official implementation to tackle longer sequences consisting of the large number of headers from Spider dataset. Please see \textbf{Supplementary File} for more implementation details.





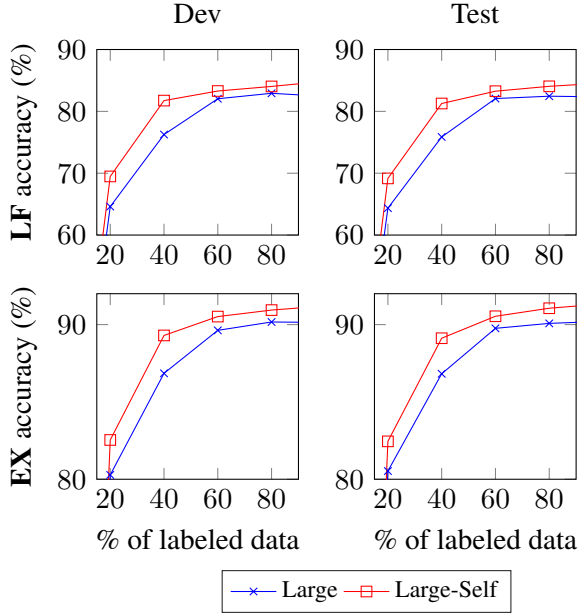
\begin {figure}
\begin{tikzpicture}
    \begin{groupplot}[
        group style={
            group name=large_lg,
            group size=2 by 1,
            ylabels at=edge left,
            xlabels at=edge bottom,
            horizontal sep=1cm,
            vertical sep=1cm,
            },
        xtick={0.0, 20, 40, 60, 80, 100},
        ytick={0, 20, 40, 60, 70, 80, 90, 100},
        ymax=90,
        ymin=60,
        xmax=90,
        xmin=15,
        width=0.55\linewidth,
        height=115.0,
	    legend style={font=\small},
        ylabel={\textbf{LF} accuracy (\%)},
        ylabel near ticks
    ]
        \nextgroupplot[title=Dev]
            \addplot[color=blue, mark=x]
                coordinates {
                (0.0,0.0)(20,64.60)(40,76.25)(60,82.04)(80,82.92)(100,82.42)
                };
            \addplot[color=red, mark=square,]
                coordinates {
                (0.0,0.0)(20.0,69.48)(40,81.74)(60,83.29)(80,84.03)(100,84.87)
                };
        \nextgroupplot[title=Test]
            \addplot[color=blue, mark=x,]
                coordinates {
                (0.0,0.0)(20.0,64.35)(40,75.86)(60,82.07)(80,82.45)(100,82.34)
                };
            \addplot[color=red, mark=square,]
                coordinates {
                (0.0,0.0)(20.0,69.16)(40,81.27)(60,83.26)(80,84.05)(100,84.57)
                };
    \end{groupplot}
\end{tikzpicture}

\begin{tikzpicture}
    \begin{groupplot}
    [
        group style={
            group name=large_ex,
            group size=2 by 1,
            xlabels at=edge bottom,
            ylabels at=edge left,
            horizontal sep=1cm,
            vertical sep=1cm,
            },
        xtick={0.0, 20, 40, 60, 80, 100},
        ytick={0, 20, 40, 60, 70, 80, 90, 100},
        ymax=92,
        ymin=80,
        xmax=90,
        xmin=15,
        width=0.55\linewidth,
        height=115.0,
        , ylabel={\textbf{EX} accuracy (\%)}
        , ylabel near ticks
        ,legend style={
            font=\mystrut
            ,draw=white!20!black
            ,legend cell align=center
            ,at={(-0.15, -0.2)},anchor=north
            }
        , xlabel={\% of labeled data}
    ]
        \nextgroupplot[]
            \addplot[color=blue, mark=x]
                coordinates {
                (0.0,0.0)(20.0,80.28)(40,86.85)(60,89.63)(80,90.17)(100,90.14)
                };
            \addplot[color=red, mark=square,]
                coordinates {
                (0.0,0.0)(20.0,82.54)(40, 89.3)(60,90.52)(80,90.94)(100,91.21)
                };
        \nextgroupplot[]
            \addplot[color=blue, mark=x,]
                coordinates {
                (0.0,0.0)(20.0,80.53)(40,86.82)(60,89.76)(80,90.08)(100,90.21)
                };
            \addplot[color=red, mark=square,]
                coordinates {
                (0.0,0.0)(20.0,82.45)(40,89.12)(60,90.54)(80,91.06)(100,91.34)
                };
    \end{groupplot}
\end{tikzpicture}

\begin{tikzpicture} 
    \begin{axis}[%
    hide axis,
    legend columns=2,
    xmin=10,
    xmax=50,
    ymin=0,
    ymax=0.4,
    legend style={draw=white!15!black,font=\small, legend cell align=center},
    ]
    \addlegendimage{blue,mark=x}
    \addlegendimage{red,mark=square}
    \addlegendentry{Large};
    \addlegendentry{Large-Self};
    \end{axis}
\end{tikzpicture}
\vspace{-1.9in}
\caption{Experimental results on \textbf{WikiSQL} with low-resource setting. y-axis is accuracy of logical-form (\textbf{LF}) or execution (\textbf{EX}).  x-axis is the amount of annotated samples for supervised learning.}
\vspace{-0.1in}
\label{figure_train_amt_wikisql}

\end{figure}

\subsection{Low-resource Environment}
In this experiment, we show the models benefit from our self-supervised method in the low-resource setting where there are few annotated training samples.
As a function of the number of annotated SQL statements, we measure the accuracy of logical form and execution with 20\%, 40\%, 60\%, 80\% of annotates SQL statements from the train split of WikiSQL. On the other hand, we use  50\% to 90\% of labeled samples for Spider dataset since it contains much smaller number of data. In order to construct the small amount of annotated training dataset, we use only the given percentages of samples from the train split and discard the rest of them.

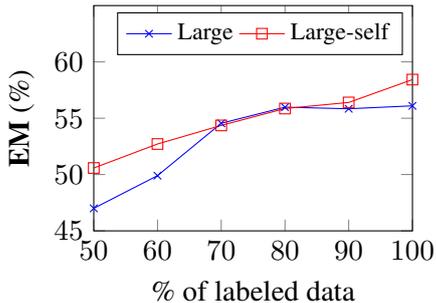
\begin{figure}
\centering
\begin{tikzpicture} 
    \begin{axis}[
    xtick={50, 60, 70, 80, 90, 100},
    ytick={45, 50, 55, 60, 70},
    ymax=65,
    ymin=45,
    xmax=100,
    xmin=50,
    width=0.75\linewidth,
    height=130.0,
    legend cell align={left},
	legend columns=2,
	legend pos= north east,
	legend style={font=\small},
    xlabel={\% of labeled data},
    ylabel={\textbf{EM} (\%)}
    , ylabel near ticks
    ]
    \addplot[color=blue, mark=x,]
        coordinates {
        (50, 47.00)(60,49.90)(70,54.55)(80,55.98)(90,55.84)(100,56.10)
        };
    \addplot[color=red, mark=square,]
        coordinates {
        (50, 50.58)(60,52.71)(70,54.36)(80,55.86)(90,56.40)(100,58.43)
        };
    \legend{Large, Large-self}
    \end{axis}
    
\end{tikzpicture}
\caption{Experimental results on \textbf{Spider} wtih low-resource setting. y-axis denotes  Exact-Set Match with value(\textbf{EM}) and  x-axis shows the percentage of labeled data from training dataset for supervised training.}
\label{figure_train_amt_spider}
\end{figure}



As shown in Figure \ref{figure_train_amt_wikisql}, Large model with our method gets the largest gains when using 40\% of training data of WikiSQL. It improves the performance of the logical form accuracy over +5\% (75.86\% $\rightarrow$ 81.27\% in the test set) and the logical-form over +2\% (86.86\% $\rightarrow$ 89.12\% in the test set). Our method also improves the performance by large margin for all the other settings. Moreover, our proposed method enables the model trained with only 60\% of labeled  data to achieve 83.26\% and 90.54\% accuracy of logical form and execution, which is better than the baseline models trained with the full labeled data.
As shown in Figure \ref{figure_train_amt_spider}, our method shows better performances with 90\% of labeled data from Spider, compared to the baseline Large. Although there are no noticeable gains at points 70\% and 80\%,  our proposed method shows large gains when there are less than 70\% of labeled data. More specifically,  the model with self-supervised training has improvement over +3\% (47.00\% $\rightarrow$ 50.58\%) with 50\% of labeled data,.

\begin{table}[t]
	\centering
	\begin{tabular}{llcc}
	
		\midrule[1pt]
	    \multicolumn{2}{c}{\textbf{SQL Clause}}&\multicolumn{2}{c}{\textbf{Accuracy (\%)}}\\
	    {}&{}&{\textbf{Large}}&{\textbf{Large-Self}}\\
		\midrule[0.8pt]
		
        \multirow{3}{*}{\rotatebox[origin=c]{90}{\textbf{WikiSQL}}}&{Select-\textbf{C}}&{97.10 / 97.01}&{97.33 / 97.20}\\
        &{Where-\textbf{C}}&{95.64 / 95.42}&{\textbf{96.83} / \textbf{96.74}}\\
        &{Where-\textbf{V}}&{96.08 / 95.77}&{\textbf{97.33} / \textbf{96.91}}\\
        &{}&{}&{}\\
		\midrule[0.8pt]
		
        \multirow{8}{*}{\rotatebox[origin=c]{90}{\textbf{Spider}}} &{Select-\textbf{C}}&{84.74}&{\textbf{86.58}}\\
        {}&{Where-\textbf{C}}&{81.19}&{81.12}\\
        {}&{Where-\textbf{V}}&{85.12}&{84.55}\\
        {}&{Group-\textbf{C}}&{69.85}&{\textbf{73.53}}\\
        {}&{Having-\textbf{C}}&{99.23}&{98.70}\\
        {}&{Having-\textbf{V}}&{96.10}&{\textbf{97.40}}\\
        {}&{Order-\textbf{C}}&{94.51}&{94.51}\\
        {}&{Order-\textbf{V}}&{91.98}&{91.98}\\
		
		\midrule[1pt]
		
	\end{tabular}
	\caption{Experimental results on SQL clauses closely related to header-span alignment task. \textbf{C} and \textbf{V} indicate the column and value predictions, respectively. Performance gaps larger than 1\% are highlighted.}
	\label{table_parts_wikisql}
\end{table}

\subsection{Fine-grained analysis}
We analyze the accuracy of each component of SQL statements in fine-grained manner. In order to validate the effectiveness of our method, we evaluate accuracy of each SQL parts that are closely related to the header-span alignment task. We report the accuracy of column and value predictions from Select / Where / Group-by / Having / Order-by clauses. Note that SQL statements from WikiSQL are composed of Select and Where clauses. The letters \textbf{C} and \textbf{V} denote column and value prediction for each part of SQL statement, respectively. As shown in Table \ref{table_parts_wikisql}, our proposed self-supervised learning framework improves the performance of all the clauses in WikiSQL. In Spider, ours outperforms Large model with large margin for certain clauses such as Select column +1.84\%, Group-by column +3.68\%, and Having value +1.30\%, although there is a slight degradation of the performance for the other clauses. 

\begin{table}
	\centering
	\resizebox{0.48\textwidth}{!}{
	\begin{tabular}{lcc}
	    \toprule[1pt]
		
		{\textbf{Model}} &{\textbf{dev (L/E)}} &{\textbf{test (L/E)}}\\
        \midrule[1pt]

		{BERT-Base \textbf{w/o}  Self} & {81.65 / 89.45} & {81.35 / 89.39}\\
		\midrule[0.4pt]
		{+Self \textbf{w.} Train} & {83.59 / 90.45} & {82.94 / 90.38}\\
		{+Self \textbf{w.} Train + Dev} & {83.13 / 90.52} & {82.44 / 90.26}\\
		{+Self \textbf{w.} Train + Test} & {82.79 / 90.45} & {82.49 / 90.53}\\
		{+Self \textbf{w.} All} & {\textbf{83.66} / \textbf{90.68}} & {\textbf{83.61} / \textbf{90.84}}\\
		
        \midrule[1.0pt]
		{BERT-Large \textbf{w/o}  Self} & {82.42 / 90.14} & {82.34 / 90.21}\\
		\midrule[0.4pt]
		{+Self \textbf{w.} Train} & {84.27 / 91.22} & {84.37 / 91.28}\\
		{+Self \textbf{w.} Train + Dev} & {84.84 / \textbf{91.44}} & {84.36 / \textbf{91.43}}\\
		{+Self \textbf{w.} Train + Test} & {84.47 / 91.32} & {84.22 / 91.37}\\
		{+Self \textbf{w.} All} & {\textbf{84.87} / 91.34} & {\textbf{84.57} / 91.21}\\
		\midrule[1pt]
		
	\end{tabular}}
	\caption{Logical-form (\textbf{L}) / execution (\textbf{E}) accuracy of self-supervised learned models with different subsets of \textbf{WikiSQL} table data.}
	\label{table_data_ablation_wikisql_}
\end{table}

\subsection{Table Data Ablation}
We show that our proposed self-supervised learning  brings significant improvements to Text-to-SQL task in the previous experiments. However, one can ask a question that where those improvements come from? We can think of two possible sources of improvements. First, with our self-supervised learning framework, the models  learns valuable knowledge of the header-span alignment task and utilize it for decoding the SQL statements. Second, the model memorizes all possible alignments between entities from column data and table headers and exploits them to predict spans of utterances during the test time. For training more reliable and scalable Text-to-SQL models, the former is much more desirable effect of self-supervised learning than the latter.

To verify the effectiveness of our proposed self-supervised learning, we design an ablation study on table data used in self-supervised training. In Table \ref{table_data_ablation_wikisql_} and Table \ref{table_data_ablation_spider_}, we report the performances of models trained with different set of unlabeled table data by discarding the tables from train, dev, or test set. If the model does not learn any useful knowledge on the header-span alignment task from the self-supervised learning and utilizes the advantage of memorizing data, the model cannot generalize to the unseen tables. As shown in the results, our method consistently improves the performance of the baseline models for both WikiSQL and Spider dataset, regardless of how we choose the subset of unlabeled table data. As expected, the more table data is available, the more there is a gain of improvement.
Based on this ablation study, we argue that Text-to-SQL model learn useful knowledge of the header alignment task from our self-supervised training, not just memorizing the contents in the tables. 

\begin{table}
	\centering
	\resizebox{0.48\textwidth}{!}{\begin{tabular}{lc}
	    \toprule[1pt]
		{\textbf{Model}} &\textbf{EM}\\
		\midrule[0.4pt]
		{Without Self (Base / Large)} & {43.90 / 56.10}\\
		\midrule[0.4pt]
		{+Self with Train} & {44.57 / 56.69}\\
		{+Self with Dev} & {44.21 / 56.78} \\
		{+Self with Train + Dev} & {\textbf{45.25 / 58.43}} \\
		\midrule[1pt]
	\end{tabular}
	}
	\caption{Exact-Set Match  with Value (\textbf{EM}) of self-supervised learned models with different subsets of \textbf{Spider} table.}
	\label{table_data_ablation_spider_}
\end{table}
\section{Conclusion and Future work}
In this work, we proposed a novel self-supervised learning method for Text-to-SQL task. Our method utilizes unlabeled table contents to train Text-to-SQL models with the header-column alignment task to learn useful knowledge of the header-span alignment. Moreover we show that model is able to learn such knowledge only with the tables from the given dataset, not resorting to the large scale external copora. We empirically validated that our method significantly improved the performance of the baseline models on several experiments. In particular, we showed that the model largely benefits from our self-supervised learning in the low resource environment where the number of human-annotated samples is small.

However, our method mainly focuses on the performance improvements of certain parts of SQL statement for the predictions of column and value, which are closely related to the header-column alignment task. As a future work, we will extend our self-supervised learning framework so that it can improve the accuracy of other parts of SQL statement.


\bibliography{main}
\bibliographystyle{acl_natbib}

\clearpage
\appendix
\section*{Appendix}
\renewcommand{\thesubsection}{\Alph{subsection}}

\subsection{Implementation Details of TaBERT}
We download the checkpoint of pretrained TaBERT \cite{yin20acl} Base and Large from the Github page \footnote{\url{https://github.com/facebookresearch/TaBERT}} and replace the backbone network of \textbf{BERT-Base} and \textbf{Bert-Large} with the pretrained TaBERT. For a fair comparison with the other baselines, we disable the snapshot mechanism which access the contents of tables during the test time.

However, we have found out that the official implementation of TaBERT cannot properly handle the long sequences consisting of the large number of headers from Spider dataset. As shown in the code snippet \ref{code}, the model outputs the embedding sequence up to the length 61 for the given sequence of which length is 65. This causes an error for finetuning the model for column prediction because the ground truth target prediction would be located after the truncated sequences. In order to tackle the issue, we pad the truncated sequence with zeros up to the length of original sequence before the truncation. 

\onecolumn
\begin{python}
# import official implementation of TaBERT
# https://github.com/facebookresearch/TaBERT
from TaBERT.table_bert import TableBertModel, Table, Column

# load TaBERT model
model = TableBertModel.from_pretrained(
        'TaBERT/weight/tabert_large_k1/model.bin')

# table example that cause a sequence length problem
# columns for each tables
columns = [['*'], ['third party companies company id', 
'third party companies company type',
'third party companies company name', 
'third party companies company address', 
'third party companies other company details'], 
['maintenance contracts maintenance contract id', 
'maintenance contracts maintenance contract company id',
'maintenance contracts contract start date', 
'maintenance contracts contract end date', 
'maintenance contracts other contract details'],
['parts part id', 'parts part name', 'parts chargeable yn',
'parts chargeable amount', 'parts other part details'], 
['skills skill id', 'skills skill code', 
'skills skill description'], 
['staff id', 'staff name', 'staff gender', 'other staff details'], 
['assets asset id', 'assets maintenance contract id', 
'assets supplier company id', 'assets asset details', 
'assets asset make', 'assets asset model', 
'assets asset acquired date', 'assets asset disposed date', 
'assets other asset details'],
['asset parts asset id', 'asset parts part id'], 
['maintenance engineers engineer id',
'maintenance engineers company id',
'maintenance engineers first name',
'maintenance engineers last name', 
'maintenance engineers other details'], 
['engineer skills engineer id', 'engineer skills skill id'], 
['fault log entry id', 'fault log asset id', 
'fault log recorded by staff id', 'fault log entry datetime', 
'fault log fault description', 'fault log other fault details'], 
['engineer visits engineer visit id', 
'engineer visits contact staff id',
'engineer visits engineer id', 
'engineer visits fault log entry id', 
'engineer visits fault status', 
'engineer visits visit start datetime',
'engineer visits visit end datetime', 
'engineer visits other visit details'], 
['part faults part fault id', 'part faults part id', 
'part faults fault short name', 'part faults fault description', 
'part faults other fault details'], 
['fault log parts fault log entry id', 
'fault log parts part fault id',
'fault log parts fault status'], 
['skills required to fix part fault id',
'skills required to fix skill id']]

# list of all columns
columns = [y for x in  l for y in x]

# how many columns are there
print(len(columns))
# 65

# construct input table structures for TaBERT
table = Table(id='', 
        header=[Column(column, 'text', sample_value='') 
                for column in columns],
        data=[]).tokenize(model.tokenizer)

# encode data by using the TaBERT model
context_encoding, column_encoding, info_dict = model.encode(
    contexts=[model.tokenizer.tokenize("dummy question")],
    tables=[table]
)

# 
print(column_encoding.size())
# (batch size, sequence length, hidden size)
# (1, 61, 1024)

# sequence mismatch between 65 and 61
\end{python}

\end{document}